\documentclass[conference]{IEEEtran}

\usepackage{amsmath,amssymb,amsfonts}
\usepackage{algorithm}
\usepackage{algorithmic}
\usepackage{graphicx}
\usepackage{textcomp}
\usepackage{ulem}
\usepackage{float}
\usepackage{booktabs}
\usepackage{hyperref,comment}   
\usepackage[table,xcdraw]{xcolor}
\usepackage{cite}

\newcommand{\add}[1]{\textcolor{violet}{#1}}

\def\BibTeX{{\rmB\kern-.05em{\sci\kern-.025emb}\kern-.08emT\kern-.1667em\lower.7ex\hbox{E}\kern-.125emX}}

\begin{document}

\title{Fast Key Points Detection and Matching for Tree-Structured Images
\thanks{The source code has been uploaded to github: \url{https://github.com/iSeND-Clemson/Fast-Key-Points-Detection-and-Matching}}
}

\author{
\IEEEauthorblockN{
Hao Wang}
\IEEEauthorblockA{\textit{School of Computing} \\
\textit{Clemson University}\\
Clemson, SC \\
hao9@clemson.edu}
\and
\IEEEauthorblockN{
Xiwen Chen}
\IEEEauthorblockA{\textit{School of Computing} \\
\textit{Clemson University}\\
Clemson, SC \\
xiwenc@clemson.edu}
\and
\IEEEauthorblockN{
Abolfazl Razi}
\IEEEauthorblockA{\textit{School of Computing} \\
\textit{Clemson University}\\
Clemson, SC \\
arazi@clemson.edu}
\and
\IEEEauthorblockN{
Rahul Amin}
\IEEEauthorblockA{\textit{Lincoln Laboratory} \\
\textit{Massachusetts Institute of Technology}\\
Lexington, MA \\
rahul.amin@ll.mit.edu}
}

\maketitle

\begin{abstract}
This paper offers a new authentication algorithm based on image matching of nano-resolution visual identifiers with tree-shaped patterns. The algorithm includes image-to-tree conversion by greedy extraction of the fractal pattern skeleton along with a custom-built graph matching algorithm that is robust against imaging artifacts such as scaling, rotation, scratch, and illumination change. The proposed algorithm is applicable to a variety of tree-structured image matching, but our focus is on \textit{dendrites}, recently-developed visual identifiers. Dendrites are entropy rich and unclonable with existing 2D and 3D printers due to their natural randomness, nano-resolution granularity, and 3D facets, making them an appropriate choice for security applications such as supply chain trace and tracking. 


The proposed algorithm improves upon graph matching with standard image descriptors. It also improves \cite{chi2020consistency}, which faces various problems when deploying in real-world applications. For instance, image inconsistency due to the camera sensor noise may cause unexpected feature extraction leading to inaccurate tree conversion and authentication failure. Also, previous tree extraction algorithms are prohibitively slow hindering their scalability to large systems. 
In this paper, we fix the current issues of \cite{chi2020consistency} and accelerate the key points extraction up to 10-times faster by implementing a new skeleton extraction method, a new key points searching algorithm, as well as an optimized key point matching algorithm. Using minimum enclosing circle and center points, make the algorithm robust to the choice of pattern shape. We show that our algorithms outperform standard key descriptors such as SIFT, FAST, and ORB in terms of accuracy (with a margin of 6\% to 10\%), while utilizing much fewer key points (about 20\% to 80\% reduction). In contrast to \cite{chi2020consistency} our algorithm handles general graphs with loop connections, therefore is applicable to a wider range of applications such as transportation map analysis, fingerprints, and retina vessel imaging. 


\end{abstract}

\begin{IEEEkeywords}
image processing, trace and tracking, key points detection, graph matching
\end{IEEEkeywords}

\section{Introduction}

\begin{figure}
  \centering
  \includegraphics[width=0.8\columnwidth]{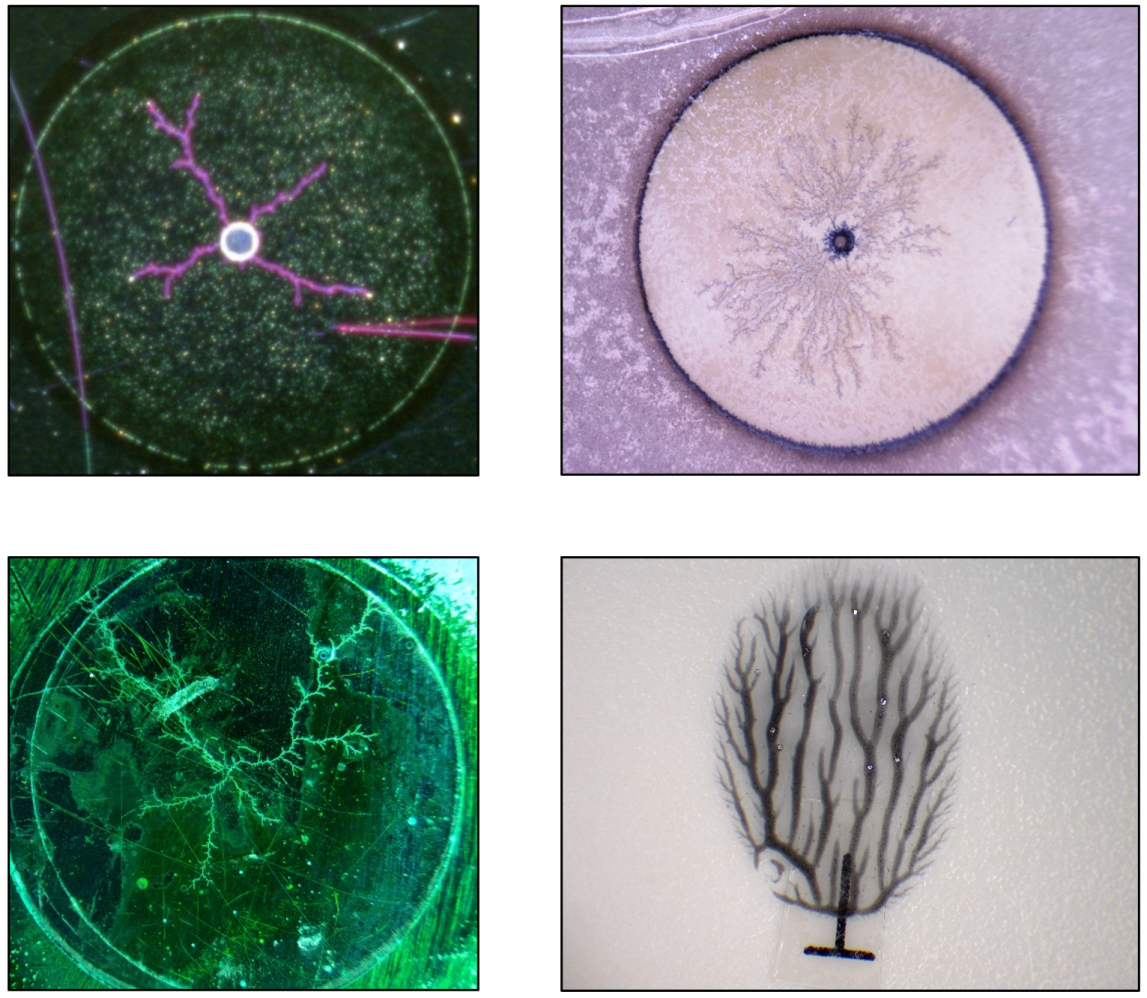}
  \caption{Dendrite samples from the laboratory.}
  \label{fig:dendrite}
\end{figure}

\begin{figure*}[ht]
    \centering
    \includegraphics[width=1\textwidth]{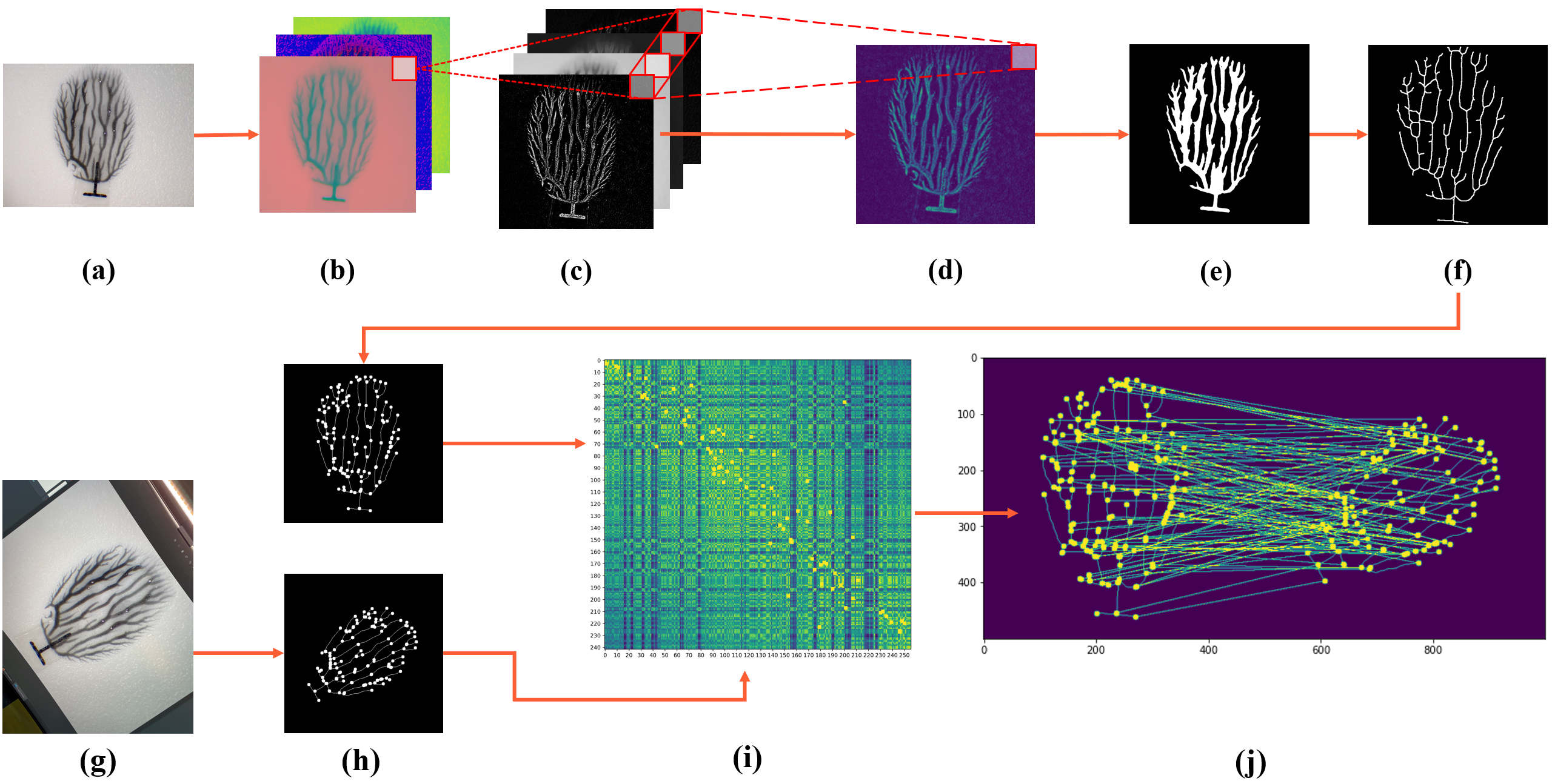}
    \caption{\textbf{Top row:} skeleton extraction of dendritic samples. Steps: (a) the raw RGB image, (b) color space expansion, (c) non-linear filtering, (d) channel fusion via PCA, (e) K-means clustering, (f) extracted single-pixel-width skeleton. \textbf{Bottom row}: Key points extraction and matching process: (g) test image taken by camera, (h) extracted key points for test and reference image, (i) similarity matrix for 200 key points using cosine and euclidean distance (brighter is more similar), (j) pairwise matching result.}
    \label{fig:ADM}
\end{figure*}

Trace and tracking are an integral part of smart connected communities due to their role in ubiquitous and uninterrupted service provisioning. Developing AI-based monitoring and tracing systems for humans, vehicles, supplies, pharmaceutical products, and objects, in general, has become one of the major key research areas in the post-Covid era \cite{kjellberg2019thinking, razi2022deep, trivedi2020digital}. For instance, tracking humans can be used by public health providers to model, analyze, and control transmissible disease \cite{mohamadou2020review, ihme2021modeling, kumari2021novel}. 

Different technologies, including biometrics, smartphones, web access points, radio signals, roadside cameras, RFID tags, contactless cards, and even vehicles' carbon footprint are used for tracking purposes \cite{trivedi2020digital}.
It is not surprising to note that most monitoring systems rely on visual data, noting that more than 90\% of information is obtained by visual perception\cite{mit2022blink}. 

A lot of identification and authentication systems for humans are nowadays based on image processing of biometrics, including facial recognition \cite{zhao2003face}, fingerprints \cite{maltoni2009handbook}, iris \cite{daugman2009iris}, ear, and palm recognition \cite{han2012palm}.
However, the use of biometrics raises privacy concerns \cite{van2003biometrics}. Secondly, these methods are specific to humans, and not usable for object tracking in general. 

Non-human tracking systems heavily rely on a variety of printed tags and web registration. For instance, barcodes and Quick Response (QR) codes supported by web-based back-end database systems are used for tracking merchandises~\cite{kjellberg2019thinking}, and store information \cite{tiwari2016introduction}. 

 More advanced electronics-based devices can be used for secure tracking. For instance, contactless media such as Radio Frequency IDentification (RFID) tags are used for authentication in facility securing, mobile payment, EZ-pass, and supply chain tracking \cite{weinstein2005rfid, juels2006rfid}. However, these solutions, in addition to their vulnerability to a variety of security attacks (e.g., power analysis, cyber attacks, and side-channel attacks), can be costly for some applications, such as daily product monitoring. This highlights the need for developing more secure low-cost visual tags \cite{krombholz2014qr}.



\textit{Dendrites} are nano-resolution fractal metallic patterns grown on various substrates through an electro-chemical process \cite{kozicki2022dendritic}. Dendrites have shown potential to be used as visual identifiers, and security key generators for authentication algorithms in numerous applications \cite{chi2020consistency, wang2015nanomaterial, kala2021contactless, thiyaneswaran2020development}. 
Their unique and high-granularity patterns provide high entropy for large-scale systems. Further, their inherent randomness, material composition, nano-scaled resolution, and 3D facet provide a key feature of unclonable that makes it almost impossible to clone these tags with any affordable existing 2D and 3D printer technology \cite{kozicki2021secure}.

As shown in Fig. \ref{fig:dendrite}, dendritic patterns can be generated in different forms, density, resolution, and granularity, controlled by the fabrication process parameters, such as the induced voltage level, temperature, and discharging direction on the electrolyte fluid \cite{kozicki2021fabrication}.

In \cite{chi2020consistency}, we proposed a novel graph-based matching algorithm to extract the 2D feature points of dendritic patterns. This method contains three parts: extraction, detection, and matching.
Unlike the conventional graph matching algorithm for complex images such as Speeded Up Robust Features (SURF), Binary Robust Invariant Scalable Keypoints (BRISK), and Oriented FAST and rotated BRIEF (ORB), our method was designed to pick feature points only from the main body of the dendritic patterns. 

The proposed method in \cite{chi2020consistency} outperforms conventional feature extraction methods in terms of computational complexity by handpicking a few feature points  from the pattern skeleton representative of the overall pattern topology instead of collecting numerous key points from the entire image. 
However, this method faces various issues when applied to real scenarios. 
First, the skeleton extraction algorithm transforms the RGB image into YCbCr color space and applies K-means clustering to separate the dendritic pattern (foreground) from the background. Unfortunately, the discontinuity of the pattern can easily result from camera noise, color shift, or occluding, which can lead to excluding part of the tree pattern. This is due to using YCbCr color space, which is sensitive to imaging conditions. To address this issue, we use multiple channels by expanding the image into multiple color spaces and applying a set of filters to exploit a richer image representation. Then, we use PCA to map the result into 2D space. Our results show that this makes the extracted skeleton less sensitive to imaging artifacts and illumination conditions.

The second improvement is on key points searching. We reached about 10 times faster by replacing the depth-first search (DFS)-like traversal method in the previous version with a parallelized breadth-first method with hand-crafted initial points around the mass center as discussed in section \ref{sec:kpsearch}.

The third issue of \cite{chi2020consistency} is the sensitivity of the graph matching algorithm to minor variations of the extracted graph topology. We observed that the extracted key points from the same image under slightly different imaging conditions could be substantially different, which can cause false point matching. In short, the graph matching method from the previous version is performed by applying the Munkres matching algorithm to the pairwise similarity scores between the test and reference graphs' nodes. In \cite{chi2020consistency}, each node is represented by a linear combination of local features with predetermined constant weights. We improved the matching rate by i) using a richer set of features for each graph node (see Table I), and ii) using learnable weights. This learning does not require a dataset and we can use only one sample with data augmentation.


Finally, we note that our method is not restricted to tree-structured patterns, but is applicable to a variety of graph-based image matching algorithms, such as transportation map analysis, fingerprints, and retina vessel imaging.

\section{Methodology}

\subsection{Feature Extraction}
The overall block diagram of the proposed method is shown in Fig. \ref{fig:ADM}. The first stage of the proposed method is channel fusion. 
According to several observations from experiments, we found that only a part of a dendritic pattern can be separated from the background using clustering algorithms under a specific color space plane. Also, the resulting patterns using different color spaces are not fully consistent. 

To extract a complete pattern, we develop a multi-channel input by transforming the image from RGB space into different color spaces, including YUV, HSV, gray-scale, etc. (Fig. \ref{fig:ADM}(b)), then applying 2D image filters such as edge detector, Laplacian filter, and Gabor filters to extract edge and line features (Fig. \ref{fig:ADM}(c)). The resulting stack contains a more rich set of information at different layers. 
Then, we apply Principal Component Analysis (PCA) to fuse the channels (Fig. \ref{fig:ADM}(c)(d)), and reduce the dimension of the extracted feature stacks to a single plane. This method can be viewed as transforming the original image into the most informative representation space using optimized non-linear projection, which has shown to be effective \cite{wang2014nonlinear}. Note that non-linearity comes from using different color spaces and 2D filters since PCA itself is a linear operator. 
This approach facilitates a low-complexity skeleton extraction which is robust to image variations compared to image segmentation using a single color-space layer (like using a grayscale version, or luminance layer in YCbCr color space).

Similar to \cite{chi2020consistency}, we remain K-means clustering for unsupervised image segmentation, i.e., extracting the pattern from the background (Fig. \ref{fig:ADM}(d)(e)). Then, we use the standard Skeletonize (scikit-image) function to turn the pattern into a single-pixel-width graph (Fig. \ref{fig:ADM}(f)).


\subsection{Key Points Searching}
\label{sec:kpsearch}

\begin{figure}[h]
  \centering
  \includegraphics[width=1\columnwidth]{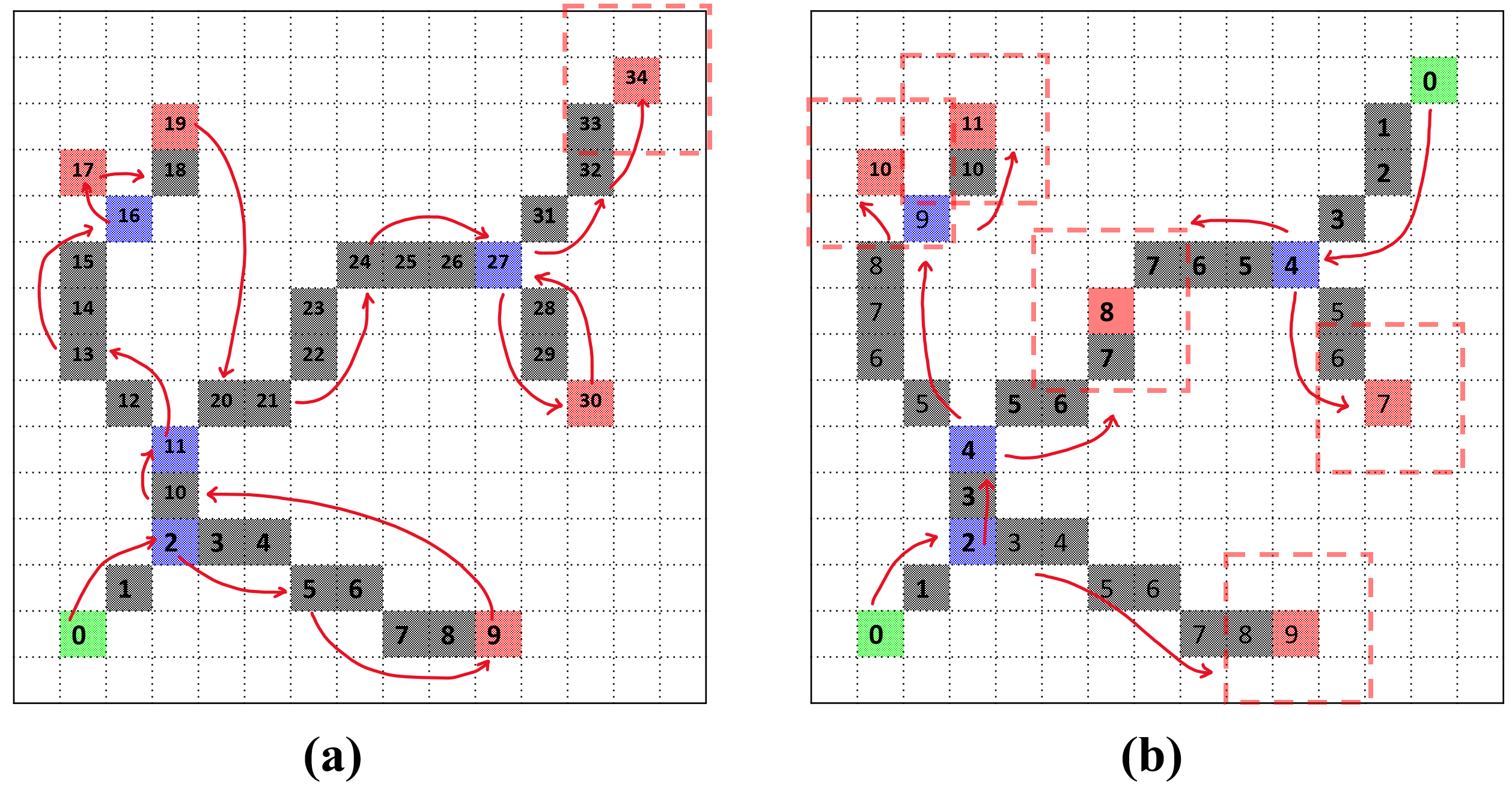}
  \caption{The search algorithm steps for a sample pattern using (a) \cite{chi2020consistency}, and (b) proposed method. Green, blue and red pixels, respectively, represent seed points, bifurcation, and endpoints. Red 3x3 windows represent the final position of the sliding windows. The proposed method requires only 11 steps compared to 34 steps required for the algorithm in \cite{chi2020consistency}. In the proposed algorithm, a sliding window is duplicated at bifurcation points and disappeared at endpoints, and traveling back is not allowed.}
  \label{fig:tree_search}
\end{figure}

Once the dendrite skeleton is extracted, we traverse through the skeleton to identify bifurcation points that form the representative tree for each dendritic pattern. In \cite{chi2020consistency}, the search starts from the center of the image and moves radially through the skeleton pixels using a depth-first-search (DFS)-like approach. This method has two drawbacks. \underline{First}, it assumes that the center of the dendrite is predefined (e.g., marked with a circle). This limits applicability to a certain class of dendrites. \underline{Secondly}, the DFS-like algorithm moves the sliding window forth and back through branches to find bifurcation points, as shown in Fig. \ref{fig:tree_search}(a). \underline{Thirdly}, the algorithm fails once it encounters a closed loop. We implement a similar method by sliding 3x3 windows through the extracted skeleton to identify bifurcation points with a few modifications to solve these issues and boost searching efficiency. 

First, we find the mass center of the dendrite pattern by simply averaging the identified key point positions. We regard the mass center as the root of the tree. If $\mathcal{N}=\{n_1,n_2,...n_N\}$ is the set of $N$ identified key points, then the mass center is $c=(\mu_x,\mu_y)=\sum_{n_i \in \mathcal{N}}{n_i}/N=\sum_{i=1}^{N}{(x_i,y_i)}/N$. Then, we find the Minimum Enclosing Circle (MEC), the smallest circular contour centered at $c$ that encompasses all nodes. 
The obtained radius would be almost half the maximum pairwise distance between the nodes of dense graphs
\begin{align}\label{eq:7}
R_{MEC}= 1/2 \max \underset{n_i, n_j \in \mathcal{N}}d(n_i, n_j)) 
\end{align}
Then, we define the initial circle, centered at $c$ with radius $R_{init}$ proportional to that of MEC ($R_{init}=\alpha R_{MEC}$ with $0 < \alpha < 1$). 
The intersection between the initial circle and the skeleton defines the set of seed points $\mathcal{S}$. 





The search algorithm starts by sliding 3x3 windows from all seed points following the skeleton branches in parallel. As these sliding windows pass through the skeleton pixels, the visited pixels are eliminated from the skeleton map. Anytime a bifurcation point is found, the sliding window is replicated, each going through an emerging branch. The window is erased once encountering an endpoint. The collected bifurcation and endpoints are recorded as key points. When two windows collide, they both disappear, and the merge point is recorded as an endpoint (Fig. \ref{fig:tree_search}(b)). The algorithm terminates when all windows disappear. 
In general, the extracted key points provide an abstract representation of the dendritic pattern retaining the overall geometrical features of the original image. 

Due to using the mass center to identify the initial circle (equivalently, seed points $\mathcal{S}$), the algorithm is robust to rotation, distortion, or any affine transformation in general. Making the initial circle radius proportional to MEC makes it scale-invariant. Eliminating visited pixels, using multiple seed points, and splitting a sliding window at bifurcation points, altogether accelerate the searching speed. Our algorithm handles cycles by merging colliding sliding windows as an additional improvement over \cite{chi2020consistency}.

\begin{table}[h]
\centering
\caption{Key point's feature information}
\label{tab:keypoint}
\resizebox{1\columnwidth}{!}{%
\begin{tabular}{lll}\toprule
\textbf{}               & \textbf{Definition}                     & \textbf{Values (Typical/Example)} \\
\\ \midrule
\textbf{X}              & Coordinate of X axis on canvas          & 0$\sim$500             \\
\textbf{Y}              & Coordinate of Y axis on canvas          & 0$\sim$500             \\
\textbf{Level}          & Level of current node                   & 0$\sim$20              \\
\textbf{Angle}          & Angle to parent node                    & -360$\sim$360          \\
\textbf{RelativeLength} & Distance to parent node                 & 0$\sim$100             \\
\textbf{Type}           & Type of node                            & root, bifurcation, end \\
\textbf{ChildIndex}     & Index of child nodes                    & {[}7,8,9{]}            \\
\textbf{LevelIndex}     & Level of child nodes                    & {[}5,10,20{]}          \\
\textbf{DistToParentX}  & Distance to X coordinate of parent node & 0$\sim$50              \\
\textbf{DistToParentY}  & Distance to Y coordinate of parent node & 0$\sim$50              \\
\textbf{ParentIndex}    & Index of parent node                    & {[}0{]}                \\
\textbf{SiblingIndex}   & Index of sibling nodes                  & {[}2,3,4{]}            \\
\textbf{DistToRoot}     & Distance to root node                   & 0$\sim$150             \\ 
\textbf{Index}          & Search order                            & 0$\sim$200             \\ \bottomrule
\end{tabular}}
\end{table}

\subsection{Graph Matching}  \label{sec:matching}

The extracted key points fully determine the representative tree for each dendrite. Therefore, the problem of identification (finding the best match between a sample dendrite pattern and a set of reference patterns in the web-based dataset) translates into evaluating similarity scores between their representative graphs (trees in this case). To assess the similarity between any two trees, we first need to find a one-to-one mapping between their points. We assign a feature vector for each keypoint that mimics local and global morphological information around the keypoint, such as type, distance to the root node (mass center), depth (number of nodes to root), orientation, parent node, sibling nodes, as well as their child node (see Table \ref{tab:keypoint} for details). This list extends the features used in \cite{chi2020consistency}. In contrast to \cite{chi2020consistency}, where each feature contributes equally to the mapping algorithm, we assign weight to features based on their contribution to the ultimate matching accuracy based on the obtained similarity score. This flexible method is driven by the fact that some features might have more essential roles in finding the right match between the nodes of two test and reference trees. Noe that we do not require a dataset for this training purpose, and it can be done with only one sample using data augmentation method.

\begin{figure}[h]
  \centering
  \includegraphics[width=0.6\columnwidth]{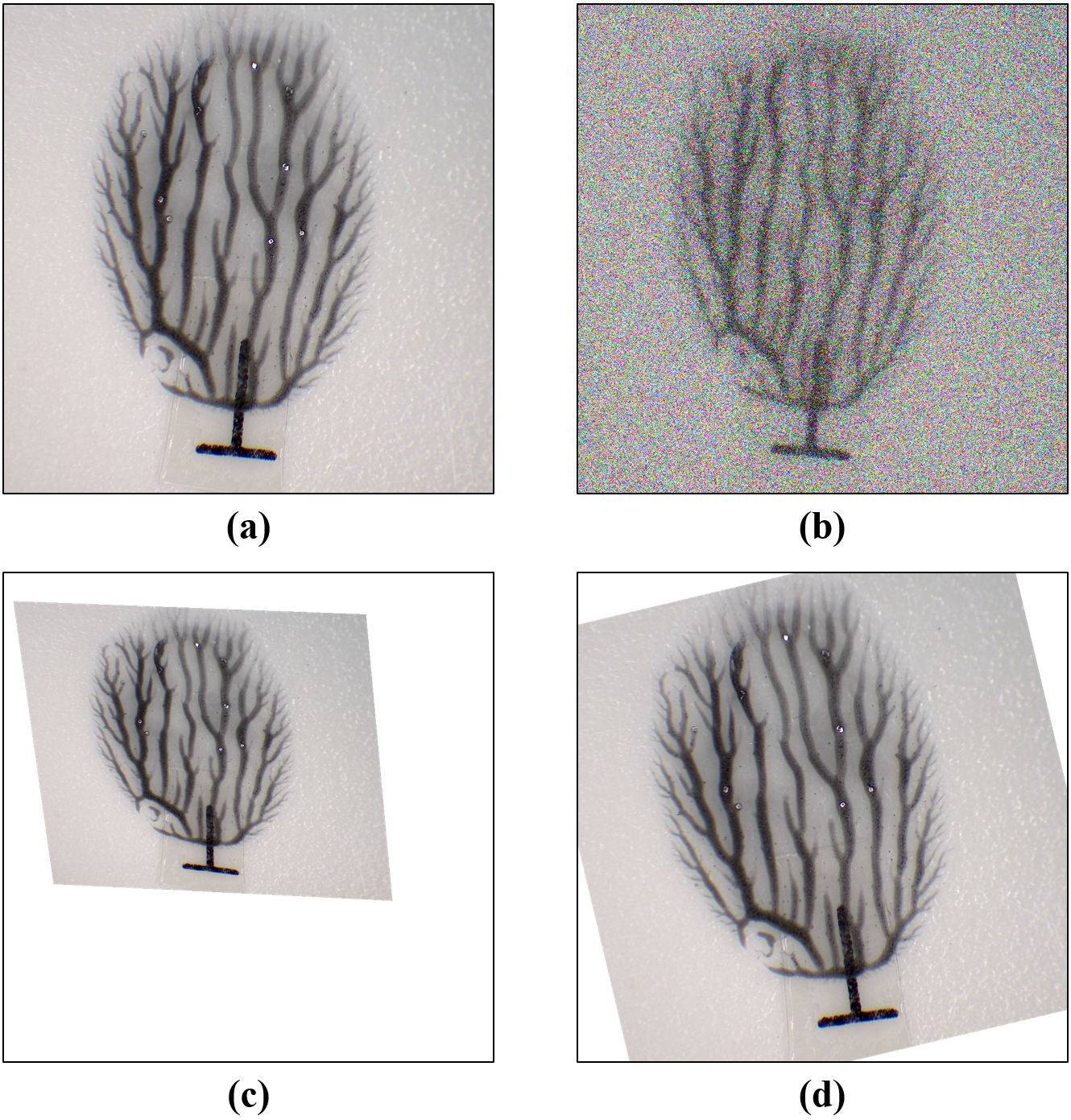}
  \caption{Samples for data augmentation, (a) raw image, (b) noisy image with Gaussian noise ($\sigma^2=0.1$), (c) image with perspective transformation (distortion degree = 0.5), (d) rotated image ($\deg = 30$).}
  \label{fig:data}
\end{figure}

To this end, we design experiments using data augmentation to evaluate the individual robustness of features against different variations. For each reference sample $R_i$, we generate a set of manipulated test samples $\{T_{i1},T_{i2},....T_{iM}\}$ by applying rotation, scaling, and perspective transformation along with noise, as shown in Fig. \ref{fig:data}.

Then we apply the proposed key points extraction method to all samples. Now, we generate distance matrices between the nodes of reference and sample trees. If $\mathbf{S}(R_i,T_{ij})=[S_{ij}]$ is the distance matrix between $R_i$ and $T_{ij}$, then $S_{kl}$ is the distance measure between node $n_k$ from tree $R_i$ and node $n_l$ from tree $T_{ij}$. In this paper, we use $d^{(F)}(n_i,n_k)=d_E^{(F)}(n_k,n_l) + d_{cos}^{(F)}(n_k,n_l)$ where $d_E^{(F)}(n_k,n_l)$ and $d_{cos}^{(F)}(n_k,n_l)$ are respectively, the normalized Euclidean and Cosine distance between nodes $n_k$ and $n_l$ using feature $F$. We normalize the measures between 0 and 1 to avoid scaling bias. Then, we apply PCA analysis on average distance metrics obtained using different features to evaluate the contribution of features in generating distance (equivalently similarity) scores. The weights obtained by PCA analysis are used to weigh features before performing the graph matching algorithm. This method provides flexibility in using an arbitrary set of features to represent key points without prior knowledge about the relevance of features. 




\begin{figure}[h]
  \centering
  \includegraphics[width=0.7\columnwidth]{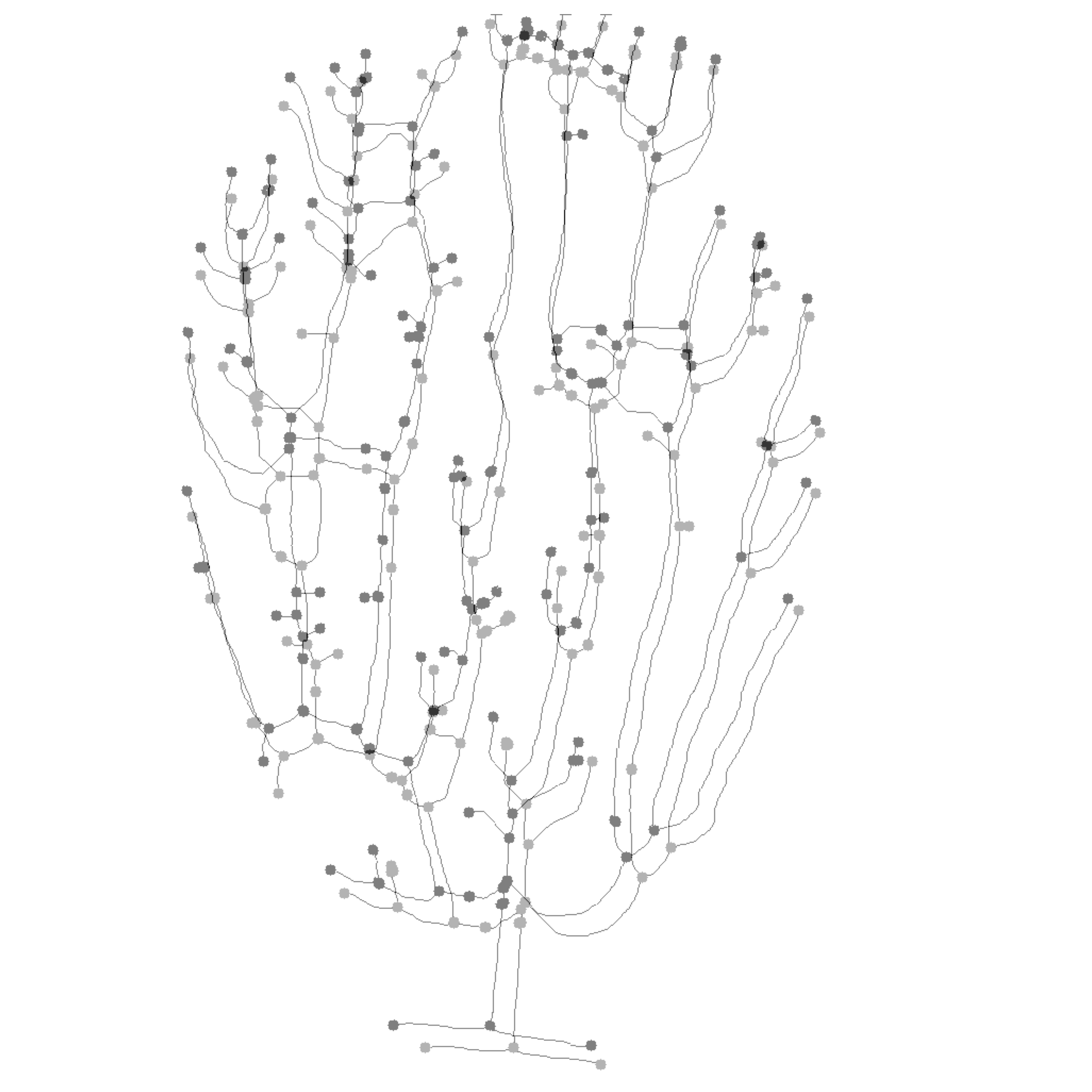}
  \caption{Using SVD to overlap two matched graphs.}
  \label{fig:svd}
\end{figure}




Then we apply Kuhn-Munkres algorithm to find the best match between the key points of the reference and test images. An exemplary distance matrix is presented in Fig. \ref{fig:ADM}(i), where pixel intensity (brighter colors) indicates higher similarity between the matched nodes.
A visualization of matching (pairing) nodes between the two graphs is provided in Fig. \ref{fig:ADM}(j). The accuracy of matching is not easily visible. In order to offer a more clear and qualitative visualization of the matching accuracy, we use SVD transformation to align the two graphs based on paired nodes.

Finally, we apply SVD to solve the transform matrix that projects corresponding $N$ key points from the test graph $T$ to the reference graph $R$. We list the sorted order of key points for the two graphs in matrices $P_{2\times N}$ and $Q_{2\times N}$. For instance, the $i^{th}$ column of $P$ is $[P_{1i}, P_{2i}]^T=[x_i-\mu_x,y_i-\mu_y]^T$ is the position of the $i^{th}$ node in the test graph ($n_i=(x_i,y_i) \in T$). 
We subtract the mean (the mass center position) $c=(\mu_x,\mu_y)$ to make it zero-mean. Likewise, the $i^{th}$ column of $Q$ is the zero-mean position of the corresponding point in the reference graph. Then, the cross-covariance matrix can be formed as
\begin{align}\label{eq:1}
    \mathbf{M} = \mathbf{P}*\mathbf{Q^T}.
\end{align} 
We compute the SVD as
\begin{align}\label{eq:2}
    \mathbf{M} = \mathbf{U}*\mathbf{\Sigma}*\mathbf{V^T},
\end{align}
It has been shown than the optimal transformation matrix that minimize the squared loss between the reference and projected points, can be obtained as \cite{sorkine2009least}: 
\begin{align}\label{eq:4}
& \mathbf{R} = \mathbf{V} * \left(\begin{array}{lllll}
1 & & & \\
& 1 & & & \\
& & \ddots & \\
& & & 1 & \\
& & & & \operatorname{det}\left(\mathbf{V} \mathbf{U}^T\right)
\end{array}\right) * \mathbf{U}^T\\
&\mathbf{p_i'} = \mathbf{c_R}+\mathbf{R}*(\mathbf{p_i-c_T}), 
\end{align}
where $X^T$ is the transpose of matrix $X$, $c_R$, and $c_T$ denotes the mass center of graphs $R$, and $T$. We expect that the transformed points of $T$ ($p_i'$) align well with the points of $R$. 
The results in Fig. \ref{fig:svd} shows a nearly-perfect alignment ($92\%$ accuracy) between the two graphs based on paired nodes using the proposed method. 

\section{Results}

\subsection{Benchmark Test}
We present results in this section to evaluate the proposed method from different perspectives, including the 
computational cost, node matching rate, and ultimate identification accuracy. Table \ref{tab:bench} 
illustrates benchmark results, where the proposed method is compared against commonly used image descriptors as well as the formerly proposed dendrite matching algorithm in \cite{chi2020consistency}. 
The overall matching loss $\mathbf{\eta}$ is defined as:
\begin{align}\label{eq:5}
    \mathbf{\eta} = (\frac{\sum_{i=1}^{N}dist(p_i',q_i)}{N}),  
\end{align}
where $dist(p,q)$ is an arbitrary distance metric, (we use Euclidean distance), N is the number of matched key points, $\mathbf{q_i}$ is the position of the $\mathbf{i}$'th key point of the reference graph, and $\mathbf{p_i'}$ is the position of the corresponding point in the test graph after performing SVD projection as detailed in section \ref{sec:matching}. 

\begin{table}[h]
\centering
\caption{COMPARATIVE ANALYSIS OF THE PROPOSED METHOD.}
\label{tab:bench}
\resizebox{0.8\columnwidth}{!}{%
\begin{tabular}{ccccc}\toprule
            & \textbf{Time} & \textbf{Key Points} & \textbf{Accuracy} \\
            \midrule
FAST                       & 0.026          & 1625                                  & 88.38             \\ 
Harris                      & 0.016          & 279                                      & 83.72             \\ 
SIFT                        & 0.091          & 527                                    & 85.58             \\ 
BRISK                       & 0.044          & 670                                      & 84.92             \\ 
ORB                         & 0.020          & 502                                      & 83.13             \\ 
Graph Matching 1 \cite{chi2020consistency}                   & 1.589          & 202                                    & 89.59             \\
\textbf{Proposed}                  & \textbf{0.131}          & \textbf{212}                                     & \textbf{95.90}             \\ \bottomrule
\end{tabular}}
\end{table}





According to Table \ref{tab:bench}, our method achieves 10-times speed compared to the reference method in \cite{chi2020consistency} due to implementing several new features such as parallel searching with multiple sliding windows, eliminating visited pixels to avoid repeated visits, and handling undesired loop scenarios in the pattern. 
Noting that both methods are slower than highly-optimized standard key descriptor methods. It is justified by the fact that our method (similar to \cite{chi2020consistency}) is customized for dendritic patterns, and includes multiple extra image processing stages to restrict the selected key points to the dendrite pattern skeleton in contrast to standard key descriptors.


The proposed method achieves a high accuracy of $95.9\%$, which is significantly higher than all competitor methods by a significant margin. The gain with respect to the closest method is over $6\%$. Note that here accuracy is the rate of successfully matched key points between the high-resolution reference image and the low-resolution and noisy test image of the same dendrite taken with a regular camera. 

Also, it is shown that an excellent matching rate of 96\% is achieved with much fewer key points compared to standard descriptors. For instance, SIFT achieves 85\% accuracy using over 500 key points, while our method achieves $96\%$ using only 212 points. The number of key points is comparable to \cite{chi2020consistency}, but our method includes 10 additional seed points. Achieving high accuracy with fewer key points reduces the computational complexity and makes it robust to imaging artifacts, noise, and scratching.

\begin{table}[h]
\centering
\caption{Comparative analysis of the proposed method under different rotation, perspective, and noise.}
\label{tab:test}
\resizebox{0.8\columnwidth}{!}{%
\begin{tabular}{cc|cc|cc} \toprule
\multicolumn{2}{c}{Rotation} & \multicolumn{2}{c}{Perspective} & \multicolumn{2}{c}{Noise} \\ \midrule
\textbf{degree} & \textbf{ACC} & \textbf{ratio}  & \textbf{ACC}  & $\mathbf{\sigma^2}$ & \textbf{ACC}               \\ \midrule
30              & 99.01      & 0.1            & 97.36          & 0.01        & 85.68       \\
70              & 98.86      & 0.2            & 94.79          & 0.1         & 79.24       \\
100             & 98.26      & 0.3            & 89.96          & 0.2         & 77.51       \\
200             & 98.21      & 0.4            & 86.91          & 0.5         & 67.47       \\
300             & 98.44      & 0.5            & 85.85          & 1           & 59.13      \\ \bottomrule
\end{tabular}%
}
\end{table}

\begin{figure}[h]
  \centering
  \includegraphics[width=0.7\columnwidth]{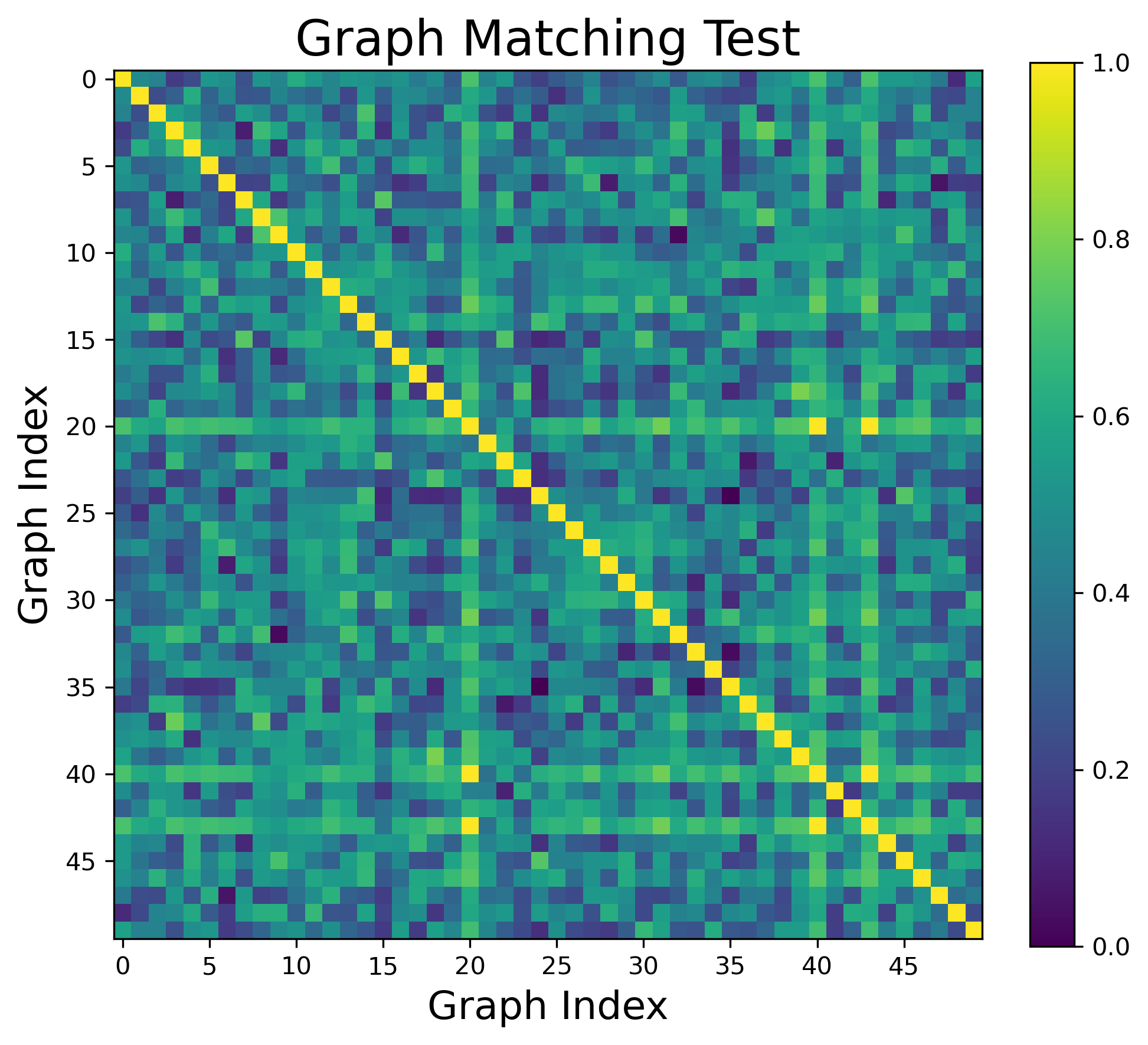}
  \caption{Graph matching test using 50 dendrite samples, pixel intensity represents the matching loss, brighter is better}
  \label{fig:kpcos}
\end{figure}

Table \ref{tab:test} demonstrates the robustness of the proposed method to rotation, perspective transformation, and noise. It is seen that the drop in matching rate accuracy is negligible under any rotation which verifies the rotation-invariance property. For perspective transformation the performance degradation remains under 12\% (from 97\% to 85\%), which is acceptable. 

For the dendrite identification test, we use 50 different dendrite samples and then apply the perspective transformation (distortion degree$=0.5$) to generate distorted test samples. In each round, we assess the similarity between the reference and all test samples. We repeat the experiment 20 times and take the average matching rate. Fig. \ref{fig:kpcos} shows the results of cross-matching of 50 different samples. The high similarity scores between the corresponding samples appear as a bright diagonal, which indicates that the proposed method approached high identification accuracy and yields excellent false-reject performance.



\begin{figure}[h]
  \centering
  \includegraphics[width=1\columnwidth]{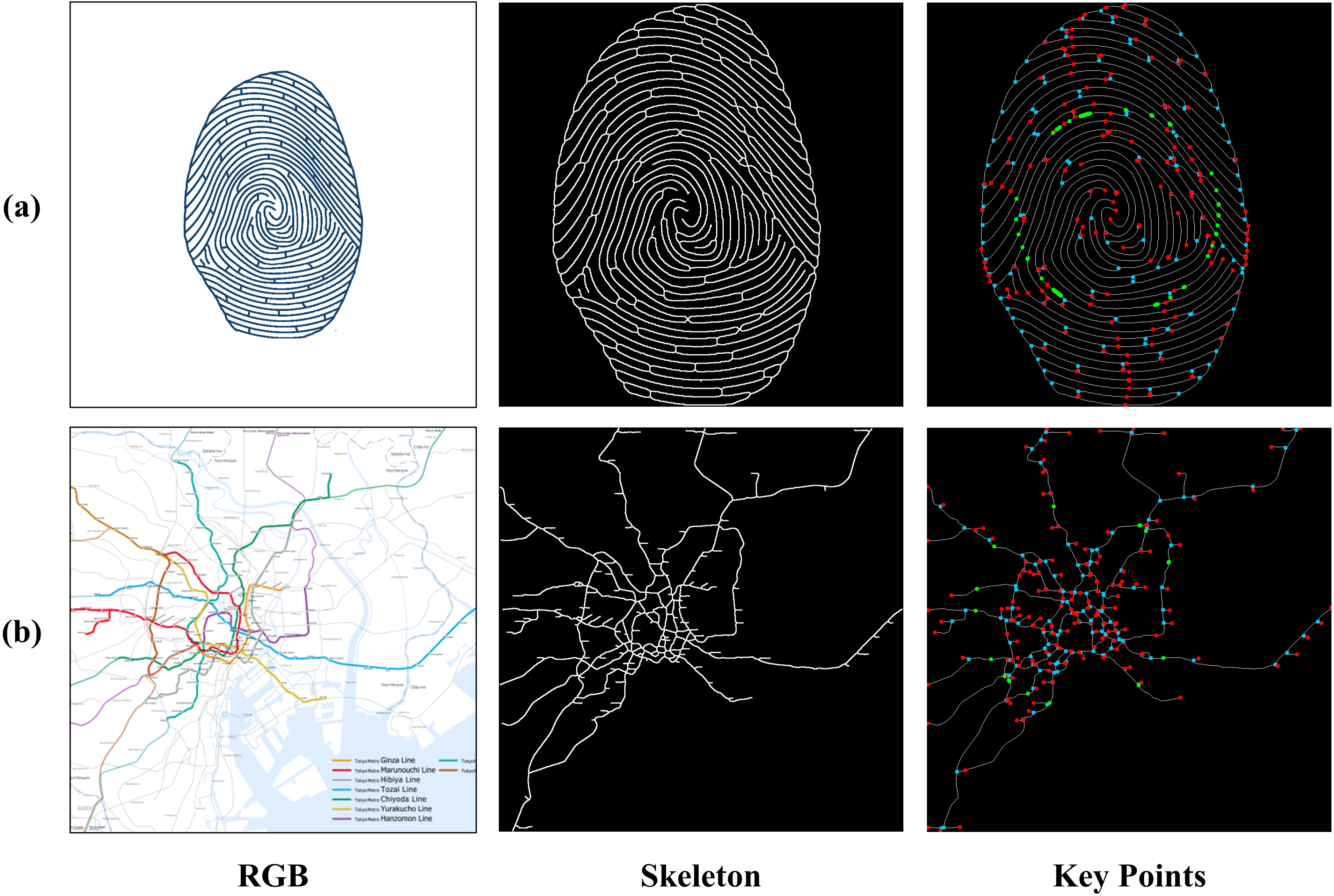}
  \caption{Application of the proposed method to other graph-structured images. Green, blue, and red pixels in the key points map, respectively, represent seed points, bifurcation, and endpoints. (a) is a fingerprint image, (b) is the map of Tokyo Metro.}
  \label{fig:tree}
\end{figure}

\subsection{Generalizability Test}

In order to demonstrate the generalizability of the proposed method beyond dendrite identification, we test the proposed method using different sample types and achieved promising results as well. 
Fig. \ref{fig:tree} presents the results for two cases, fingerprints and maps. The extracted skeleton and the extracted key points for both cases represent the overall geometry of the original image that preserves maximal information.
In the Tokyo Metro Map, benefited from the feature fusion, the proposed method has divided the subway lines (colored) from other crossed city highways. Since the metro stations and intersections appear as key points, the resulting key points can be used for metro complexity evaluation or transportation analysis \cite{stott2010automatic, chen2022network}.
Also, it is seen that in contrast to \cite{chi2020consistency}, which is applicable only to tree-structured loop-free graphs, our method is applicable to more complex graphs.




\begin{figure}[h]
  \centering
  \includegraphics[width=1\columnwidth]{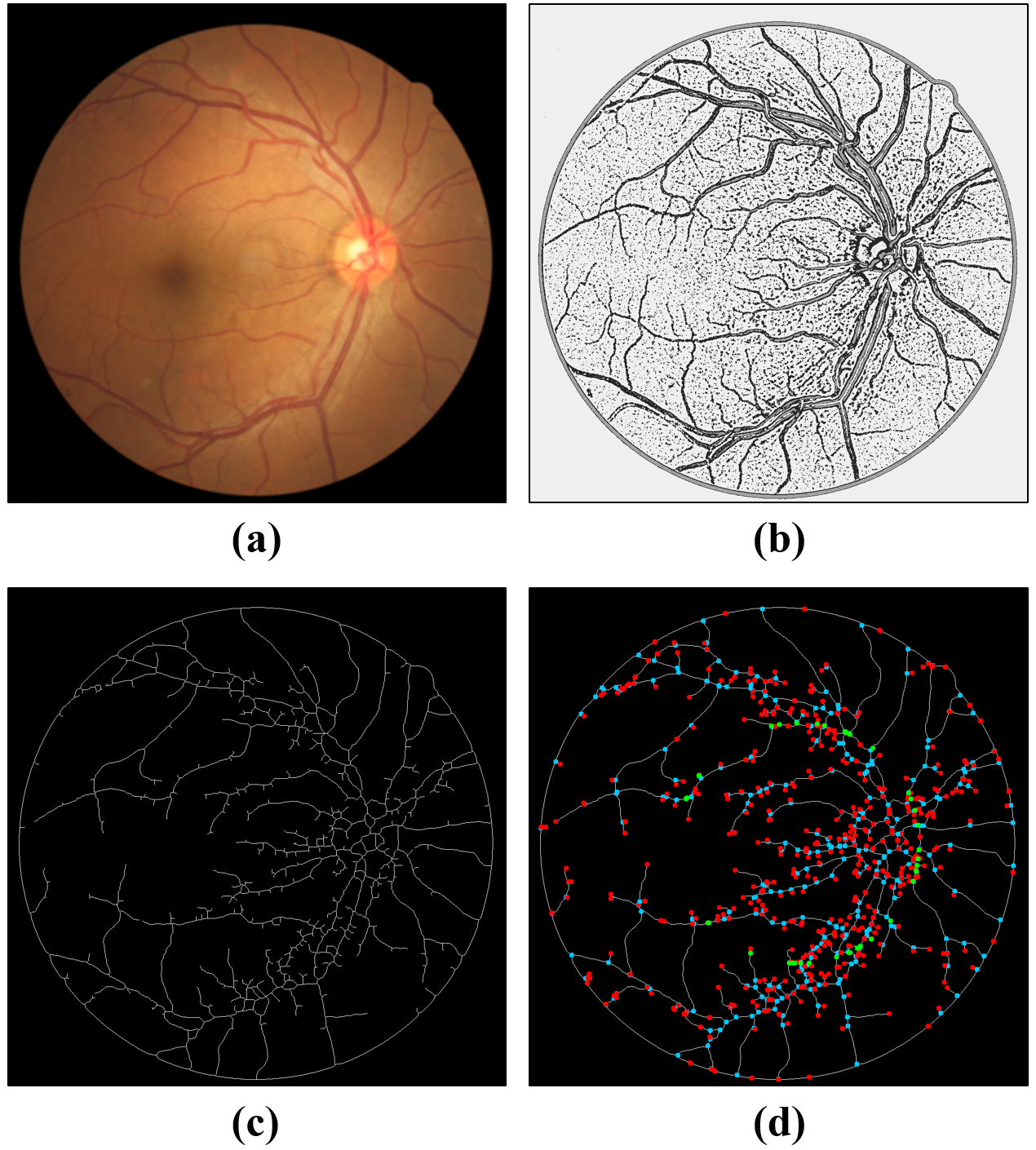}
  \caption{Application of the proposed method to Retinal Vessel images. (a) original image, (b) foreground using feature fusion and kmeans segmentation, (c) extracted skeleton, (d) key points map, green, blue, and red pixels, respectively, represent seed points, bifurcation, and endpoints.}
  \label{fig:retina}
\end{figure}

We have also tested the proposed methods on retinal vessel images. 
Retinal vessel segmentation has remained a hot research topic over the last decade. 
The information regarding age-related macular degeneration and diabetic retinopathy (DR) can be extracted from fundus screening \cite{lyu2022fractal}.
With simple modifications in the modeling parameters (such as using different color spaces and filters in channel fusion and the number of clusters (K) in the segmentation stage), the proposed method is applicable to retina images and yields reasonable performance. 

Recently, the deep learning (DL)-based methods have been applied to image segmentation, skeleton extraction, and image matching tasks \cite{fu2016deepvessel, guo2021mes, lyu2022fractal, nguyen2013effective, orlando2016discriminatively, shah2019unsupervised}. 
Although the overall performance of DL methods is much higher, our method is unsupervised and train-free that still remains advantages in specific tasks. More importantly, our proposed method is fully compatible with their results, which means we can directly utilize their extracted skeleton for key points identification and graph matching. For the application at hand, using dendrites as visual identifiers, the sharing of images may raise security and privacy concerns. Therefore, untrained methods are highly desired. 


\section{Conclusion}
We develop a new method for visual identification based on graph matching of tree-structured images. The essence of our method is extracting representative graphs from test and reference images using a channel fusion method along with a fast key points extractor. The resulting key points are used by pairwise point matching with a learnable loss function to assess the structural similarity between the two graphs. 
The experiments exhibit our proposed method has competitive performance against several benchmarks in terms of both robustness and efficiency while using much fewer key points to assess structural similarity. We improved upon the recently developed algorithm in \cite{chi2020consistency} in terms of complexity and matching rate by introducing a few novel features. Specifically, our method utilizes about 20\% to 80\% fewer key points than standard descriptors such as SIFT, BRISK, ORB, etc. Our method is about 10-times faster with respect to \cite{chi2020consistency}. More importantly, our method handles general graphs with loopy connections, extending the applications beyond tree-structured images. Therefore, it is applicable to fingerprint detection, map analysis, retinal vessel imaging, and similar tasks. Finally, note that our method is unsupervised and does not require a training dataset as opposed to deep learning-based methods, which makes it appropriate for security-restricted and privacy-preserving applications.




\bibliography{ref}
\bibliographystyle{vancouver}

\end{document}